\documentclass[10pt,twocolumn,letterpaper]{article}

\usepackage{cvpr}              

\usepackage{graphicx}
\usepackage{amsmath}
\usepackage{amssymb}
\usepackage{booktabs}
\usepackage{mathrsfs}

\usepackage[pagebackref,breaklinks,colorlinks]{hyperref}

\usepackage[capitalize]{cleveref}
\crefname{section}{Sec.}{Secs.}
\Crefname{section}{Section}{Sections}
\Crefname{table}{Table}{Tables}
\crefname{table}{Tab.}{Tabs.}

\def\cvprPaperID{3455} 
\def\confName{ICCV}
\def\confYear{2023}

\begin{document}

\title{Exploiting the Textual Potential from Vision-Language Pre-training for Text-based Person Search}

\author{
Guanshuo Wang$^{1}$, 
Fufu Yu$^{1}$, 
Junjie Li$^{1,2}$, 
Qiong Jia$^{1}$, 
Shouhong Ding$^{1}$\\
$^{1}$Tencent YouTu Lab \quad $^{2}$AI Institute, Shanghai Jiao Tong University \\
{\tt\small \{mediswang,fufuyu\}@tencent.com} \\
}

\maketitle

\begin{abstract}
Text-based Person Search (TPS), is targeted on retrieving pedestrians to match text descriptions instead of query images. 
Recent Vision-Language Pre-training (VLP) models can bring transferable knowledge to downstream TPS tasks, resulting in more efficient performance gains. 
However, existing TPS methods improved by VLP only utilize pre-trained visual encoders, neglecting the corresponding textual representation and breaking the significant modality alignment learned from large-scale pre-training.
In this paper, we explore the full utilization of textual potential from VLP in TPS tasks.
We build on the proposed VLP-TPS baseline model, which is the first TPS model with both pre-trained modalities. 
We propose the Multi-Integrity Description Constraints (MIDC) to enhance the robustness of the textual modality by incorporating different components of fine-grained corpus during training.
Inspired by the prompt approach for zero-shot classification with VLP models, we propose the Dynamic Attribute Prompt (DAP) to provide a unified corpus of fine-grained attributes as language hints for the image modality.
Extensive experiments show that our proposed TPS framework achieves state-of-the-art performance, exceeding the previous best method by a margin. 
\end{abstract}

\section{Introduction}
\begin{figure}[t]
\centering
\includegraphics[width=1.0\linewidth]{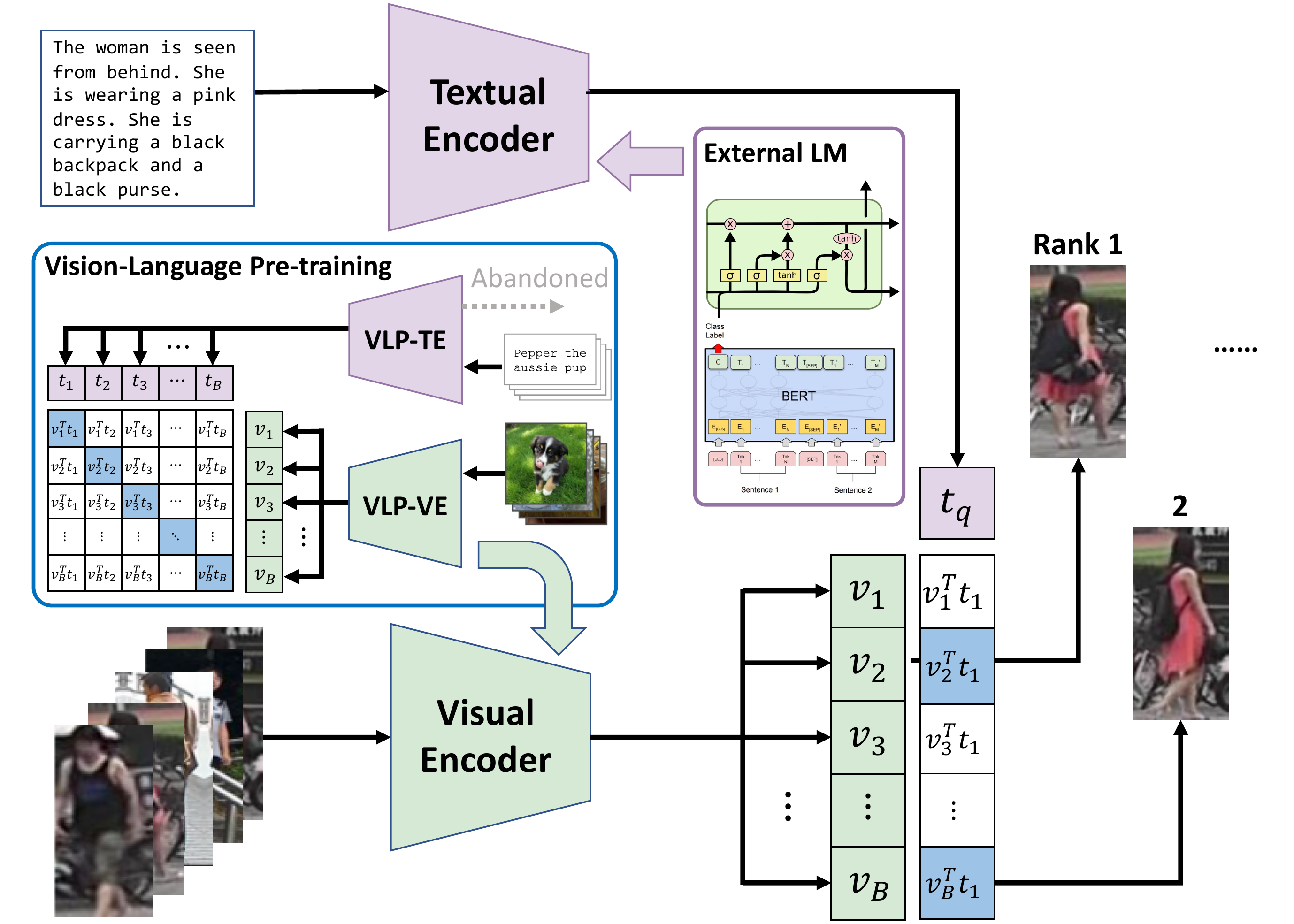}

\caption{\label{figure1} A typical Text-based Person Search model initialized with Vision-Language Pre-training models. The pre-trained vision encoders are used to initialize the TPS image representation, but textual encoders are altered by external language models such as LSTM or BERT, which is a asymmetry setting that might break the pre-trained cross-modal alignment. }
\end{figure}

Text-based Person Search (TPS) is a task focused on retrieving persons based on textual descriptions rather than query images. Figure \ref{figure1} illustrates a typical TPS pipeline, which consists of dual cross-modal backbones for visual and textual encoders. Query description texts and gallery person images are encoded into discriminative features, and retrieval ranking is determined based on modality-wise feature similarities. The training target of the dual-modality model is to maximize the similarity between visual and textual features, also known as cross-modal semantic alignment. Existing TPS methods focus on improving the modality alignment with ID discrimination through two main approaches. First, cross-modal loss functions for classification or metric targets are applied to impose explicit constraints on modality alignment, in addition to the generally used loss functions in image-level reID tasks \cite{zhang2018deep,aggarwal2020text,wang2020vitaa}. Second, local representations are designed for both visual and textual modalities to encode fine-grained details, such as partial cloth regions or pixels on images \cite{chen2018improving,ding2021semantically,han2021text,yan2022image,chen2022tipcb}. Informative fine-grained appearance details, such as the partial cloth regions or pixels on images \cite{wang2020vitaa} and annotated attributes for persons \cite{aggarwal2020text}, are also used. While most state-of-the-art methods focus on the latter approach with complex mechanisms or side information, the performance gains are limited by the model's generalization ability.

Vision-Language Pre-training (VLP) models, such as CLIP \cite{radford2021learning}, ALIGN \cite{jia2021scaling} and BEiT-3 \cite{wang2022image}, have recently emerged as powerful tools for achieving strong generalization on downstream image-text tasks. These models learn from large-scale unlabeled data of image-text pairs \cite{mokady2021clipcap,song2022clip,shen2021much}, and perform contrastive pre-training between corresponding pairwise vision and textual encoded features. While just initializing with the visual encoder of CLIP has been shown to greatly improve TPS under limited pairs of pedestrian image-text data \cite{han2021text,yan2022clip}, existing methods typically abandon the pre-trained textual encoder and alter it with an LSTM or BERT language model during fine-tuning to decrease the risk of model collapse. However, the absence of VLP textual representations may break valuable cross-modal alignment by both encoders. Existing methods attempt to re-establish this cross-modal tie based on limited data fine-tuning, which is not an optimal setting. Additionally, without the support of VLP textual encoders, existing methods cannot fully leverage the flexible nature of text modalities to enhance the alignment performance of image modalities, such as constructing training prompts to enhance the visual encoder.


In this paper, we propose to exploit the dual generalization of VLP models, particularly their textual potential, to achieve a significant performance improvement over existing models. We introduce the Vision-Language Pre-trained Text-based Person Search (VLP-TPS) model as our baseline, which effectively utilizes both pre-trained encoders and reduces training difficulty through dual-modal token pooling and low-level representation freezing for the textual modality. Leveraging pre-trained textual transferable knowledge, we propose two strategies that operate on the description corpus. To ensure various text description integrity, we introduce the Multi-Integrity Description Constraints (MIDC), which rearrange fine-grained information in attribute phrases to generate correct but different kinds of broken or additional sentences. We employ multiple constraint losses between all the original or generated textual features and image features. Inspired by the prompt method used for VLP-based zero-shot classification, we propose the Dynamic Attribute Prompts (DAP) strategy. We construct unified attribute descriptions based on type-related prompt templates according to the core attributive nouns. The dynamic prompt textual features are targeted to enhance fine-grained representations in visual features during cross-modal alignment.


The contribution of this paper can be summarized as follows: Firstly, we propose the VLP-TPS baseline, which is the first model to completely and effectively utilize dual-tower VLP vision and textual encoders. Secondly, we introduce the MIDC and DAP modules, which aim to exploit the textual potential by enhancing textual consistency and fine-grained image representations reversely guided by textual prompts. Thirdly, extensive experiments demonstrate that VLP-TPS with textual potential, denoted as TP-TPS, achieves state-of-the-art performance and outperforms existing methods by a large margin.


\begin{figure*}
\centering
\includegraphics[width=1.0\linewidth]{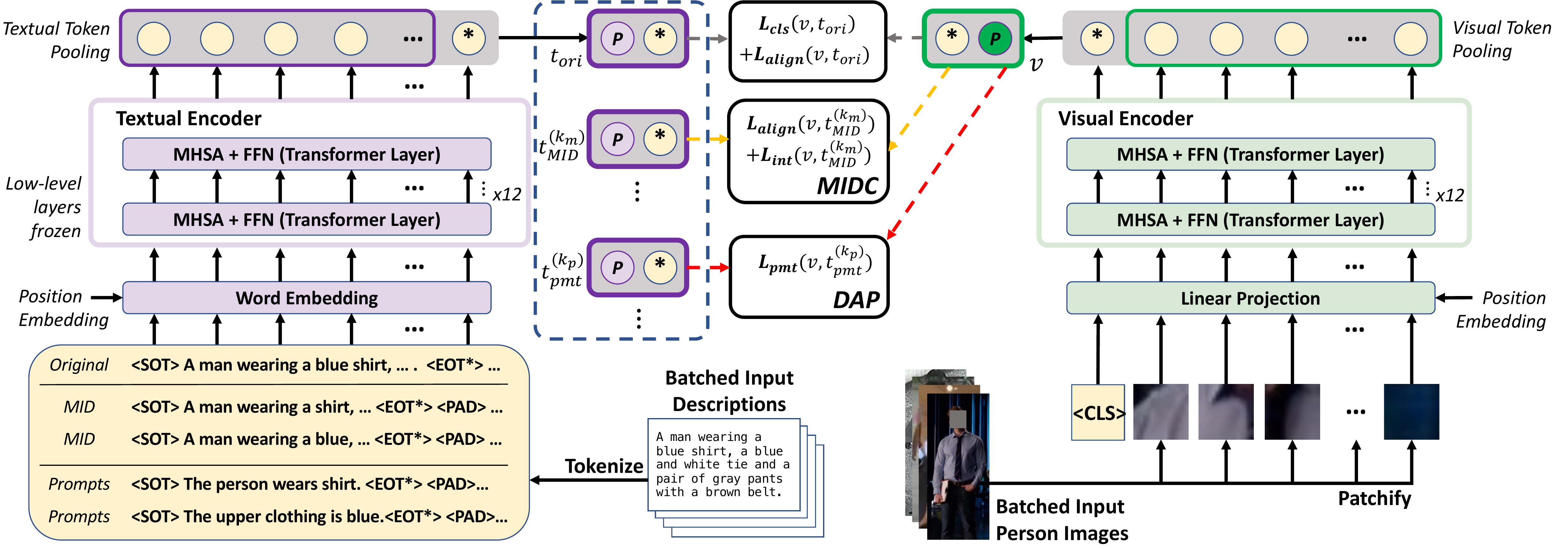}
\caption{\label{figure2} Overview pipeline of the proposed TP-TPS. A simple baseline framework is developed based on CLIP pre-trained model. Visual and textual token pooling operations are employed to represent token-level fine-grained features for both modalities. We further introduce the Multi-Integrity Description Constraint (MIDC) and Dynamic Attribute Prompt between visual features and different kinds of textual features. }
\end{figure*}

\section{Related Work}
\subsection{Text-based Person Search}
Text-based Person Search (TPS) task is firstly defined in \cite{li2017person}, also with the first benchmark CUHK-PEDES, which utilizes the visual backbone of ResNet50 network and the textual backbone of LSTM network with Word2Vec embedding to learn cross-modal representations. Some later works follow this basic setting and introduce improvements in various aspects. Zhang \textit{et al.} \cite{zhang2018deep} proposes cross-modal projection losses for classification and matching tasks (CMPC \& CMPM). Major amount of works focus on the joint representations of global and local information in cross-modal retrieval. Chen \textit{et al.} \cite{chen2018improving} first proposes to mine the local information on both visual and textual modalities. Aggarwal \textit{et al.} \cite{aggarwal2020text} introduces attribute prediction to preserve the semantics in learned features. Wang \textit{et al.} \cite{wang2020vitaa} utilizes auxiliary human parsing information for attribute-aligning. Ding \textit{et al.} \cite{ding2021semantically} establishes correspondences between body parts and noun phrases with the attention mechanism. \cite{yan2022image} employs image-specific suppression to balance the information between modalities and conducts implicit local alignment. 

TPS is benefited from the blossom of Transformer-based language models \cite{devlin2018bert, liu2019roberta}. NAFS \cite{gao2021contextual} first introduces BERT \cite{devlin2018bert} as the textual encoder of this task, combined with pyramidal image representations with non-local aggregation. TIPCB \cite{chen2022tipcb} utilizes BERT embedding as local textual representations and matches with corresponding part-based visual representations. LGUR \cite{shao2022learning} with the BERT textual encoder learns granularity-unified representations to bridge the inter-modal gap. SRCF \cite{suo2022simple} simply utilizes ResNet50 and BERT models for visual and textual feature extraction, and introduces correlation filtering to adaptively align cross-modal features with light computation burden. For the visual encoder backbone, Vision Transformer (ViT), has also been applied in some methods \cite{shu2022see,li2022learning,yan2022clip}. Some works applied image modal pre-training of CLIP \cite{han2021text,yan2022clip}, and achieves significant improvement, but alters the textual encoder with LSTM or BERT, which breaks the valuable cross-modal alignment in pre-training.

In our work, we reap the benefits from Transformer-based encoders, and try to exploit the dual generalization of VLP models. Especially we retrieve the pre-trained textual encoder to maintain the cross-modal transferable knowledge, so that we can introduce textual strategies, Multi-Integrity Description Constraints (MIDC) and Dynamic Attribute Prompts (DAP), to improve TPS performance by these textual knowledge.

\subsection{Vision-Language Pre-training}
Recently, self-supervised learning methods make pre-training models available on the large-scale unannotated database. The easier access of huge amounts of image-text pair data items without specific classification annotations urges the pre-training methods with cross-modality interaction. CLIP \cite{radford2021learning} jointly pre-trains visual and textual models by predicting correct image-text pairings with contrastive learning, and gets an efficient zero-shot transferring performance. ALIGN \cite{jia2021scaling} optimizes both image-to-text and text-to-image classification pretext tasks. FILIP \cite{yao2021filip} introduces late interaction between modalities to the contrastive learning target for better fine-grained generalization. BLIP \cite{li2022blip} combines vision-language understanding and generation tasks into a unified VLP framework. VL-BEiT \cite{wang2022image} and BEiT-3 methods \cite{bao2022vl} focus on unifying both image and language modalities pre-training to one equal masked modeling framework. Some TPS methods \cite{han2021text,shu2022see,yan2022clip} have exploited the benefits of VLP on large-scale databases. However, most previous methods only utilize the pre-trained visual encoders for better basic performance, but ignore the pre-trained language model with strong modality association with the visual. Our proposed method utilizes the CLIP model to initialize both visual and textual backbones. We also utilize the power of natural language descriptions during fine-tuning, by constraints of various integrity and prompts for attribute hints. 

\section{Method}
Our method is based on a dual-encoder backbone baseline on VLP model. Taken CLIP model as an example, we first specify a simple baseline VLP-TPS based with a targeted fine-tuning strategy. To further exploit the textual potential based on dual generalization, we propose some operations over natural languages, and combined into the total pipeline TP-TPS. The overview structure of our proposed method is illustrated in Figure \ref{figure2}.  The Multi-Integrity Descriptions (MID) are generated for constraints of cross-modal alignment. Besides, a Dynamic Attribute Prompt (DAP) module is applied for supervision to provide attribute hints from natural language descriptions. 

\subsection{TPS Baseline based on VLP Models}
As shown in Figure \ref{figure2}, the backbone network consists of independent visual and textual encoders. Person images and text descriptions are respectively fed into the corresponding encoders for the vision and textual features outputs. We follow the CLIP-ViT setting of OpenAI's release \cite{radford2021learning} to utilize the transferable knowledge of pre-training. The visual encoder is a standard Vision Transformer \cite{dosovitskiy2020image}. The input image of size $H \times W$ is first divided into equivalent $\frac{H \times W}{P^2}$ patches as input sequences, with a class token attached to the head. The fed image patches sequences are represented by an output token sequence with a length of $\frac{HW}{P^2}+1$, consists of an image-level feature corresponding to the class token, and patch-level features $\mathbf{v}=\{v^g, v^p_1, \cdots, v^p_{\frac{HW}{P^2}}\}$. 

The textual encoder is a stacked network with $L$ Transformer blocks. The input text is first tokenized into 512-dim word embeddings. The text embedding sequences are padded into the max length $L_M$ for batched inputs. We take the output token corresponding to the end of text without padding as the sentence-level feature, and $L_M$ word-level features $\mathbf{t}=\{t^g, t^w_1, \cdots, t^w_{L_M}\}$.

\begin{figure}[t]
\centering
\includegraphics[width=1\linewidth]{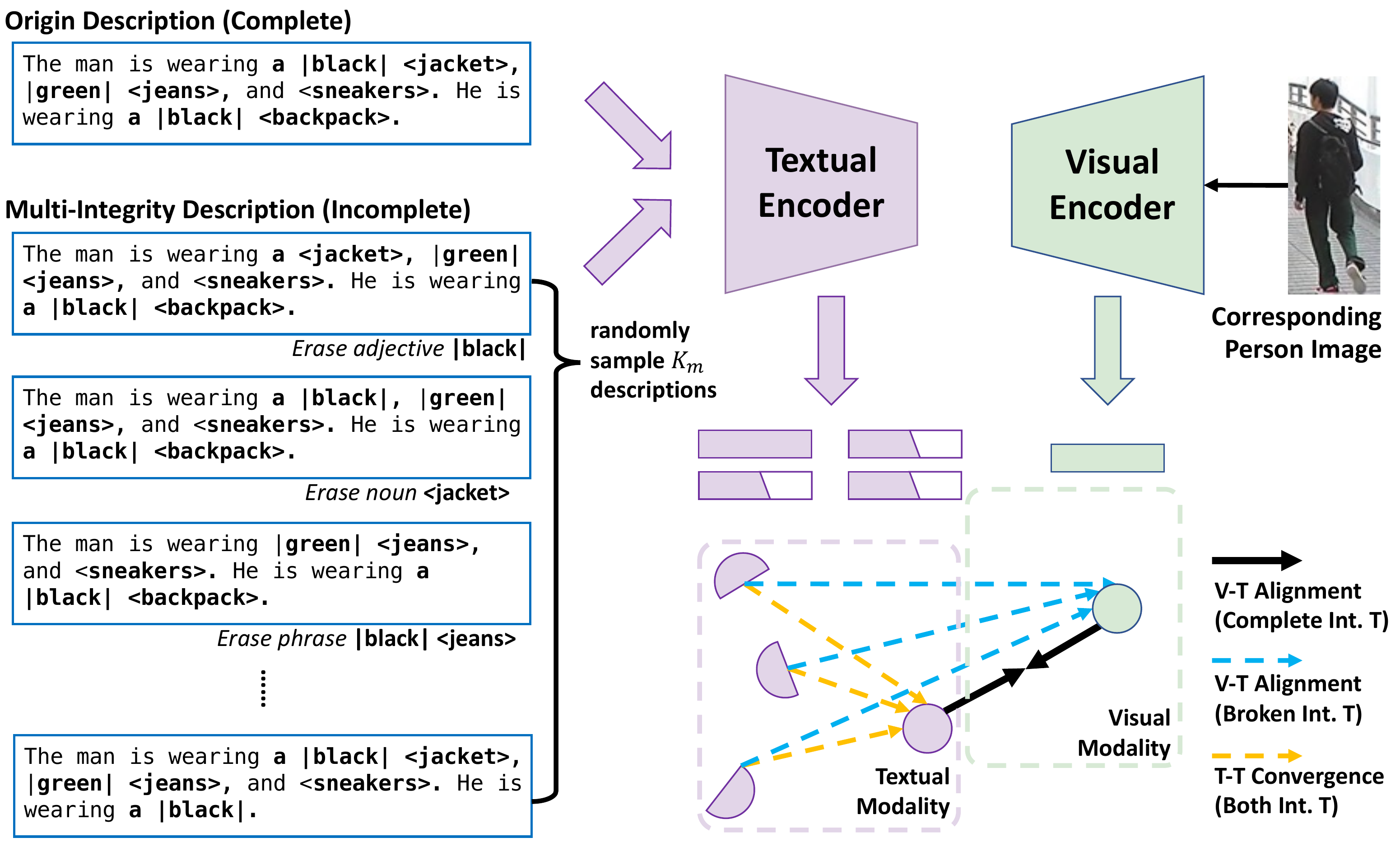}
\caption{\label{figure3} Illustration of Multi-Integrity Description Constraints. The incomplete rectangle and circles denote the features of MID with incomplete integrity. }
\end{figure}

Fine-grained information dominates the discrimination capacity for TPS tasks. Considering the mechanism of self-attention in Transformer layers, image- and sentence-level features from a specific token tend to represent ID-discriminative information jointly by weighted summation of attention maps between tokens, and dilute some minor details. Comprehensive token-wise features can include such information, so we introduce a simple token pooling on patch- and word-level features, by utilizing the Generalized Mean Pooling \cite{radenovic2018fine} as
\begin{equation}
v^P = (\frac{P^2}{HW}\sum_k{(v^p_k})^q)^{\frac{1}{q}}, \quad t^P = (\frac{1}{L}\sum_k{(t^w_k)^q})^{\frac{1}{q}} \\
\end{equation}
where $q > 1$ is a parameter, $L \le L_m$ is the actual length of input texts. This is a harmonic approach between global average and max pooling methods. We take the concatenation of global- and token-level features $v=\{v^g, v^P\}$ and $t=\{t^g, t^P\}$ for the final feature output.

It is somehow difficult to obtain a satisfactory result by directly fine-tuning the CLIP pre-trained model on TPS datasets, so several previous methods choose to alter the textual encoder with external language models. We find two reasons for this issue. On the one hand, the learning rate schedule and weight decay regularization  of TPS fine-tuning on VLP models is quite finical. On the other hand, updating all weights of the textual encoder can cause serious over-fitting, and low-level representations is somehow universal in different language tasks, so we choose to freeze several bottom layers in the textual encoder.

\subsection{Multi-Integrity Description Constraints}
During the learning language-vision representations for TPS tasks, relatively complete textual description are fed without any integrity variations. However, given descriptions are often just a few sketch words, far from comprehensive details of targeted persons. This conflict makes TPS severely relies on the quality of query text descriptions. From this view, we propose the constraints of fine-grained multi-integrity descriptions to relieve this issue.

As shown in Figure \ref{figure3}, the most valuable discriminative information in an ordinary text description $T$ is usually some attribute phrases containing adjectives and object nouns. If we manually erase any component in phrases, or erase the whole phrase, the integrity of this description sentence will be ruined as a whole description, but the correctness of reserved subsets is not affected. We can generate the Multi-Integrity Descriptions (MID) from all the possible combinations of incomplete descriptions $\mathcal{T}_{MID}=\{T^{(0)}, T^{(1)}, \cdots\}$ by erasing fine-grained components in discriminative phrases. We randomly pick up $K_m$ descriptions from $\mathcal{T}_{MID}$, and fed them into the textual encoder with the original description $T$, to get the output of complete textual feature $t$ and $K_m$ multi-integrity textual features $t^{(k)}$. 

We employ the constraint on these multi-integrity features based on two metric relationships. The cross-modal metric of cosine similarity is calculated by $S(v,t) = \frac{v^Tt}{\|v\|\|t\|}$. On the one hand, all the complete and partial textual features $t$ and $t^{(k)}$ should be aligned with their corresponding visual feature $v$. The L2-normalized softmax loss \cite{wang2017normface} is used for the classification task as
\begin{equation}
\begin{aligned}
L_{cls} = &-\sum_{i=1}^B{\log{\frac{e^{s\mathbf{W}_v^{y}\hat{v}_i}}{\sum_{c=0}^{C-1}{e^{s\mathbf{W}_v^{c}\hat{v}_i}}}}}-\sum_{i=1}^B{\log{\frac{e^{s\mathbf{W}_t^{y}\hat{t_i}}}{\sum_{c=0}^{C-1}{e^{s\mathbf{W}_t^{c}\hat{t_i}}}}}} \\
&-\sum_{i=1}^{B}\sum_{k=0}^{K_m-1}{\log{\frac{e^{s\mathbf{W}_t^{y}\hat{t}_{i}^{(k)}}}{\sum_{c=0}^{C-1}{e^{s\mathbf{W}_t^{c}\hat{t}_{i}^{(k)}}}}}},
\end{aligned}
\end{equation}
where the $\hat{v_i}$ and $\hat{t_i}$ are normalized visual and textual features, and $s$ is a scale parameter for divergence. This classification loss consists of visual, global textual and partial textual parts. Besides, we employ the alignment loss for a metric learning task as
\begin{equation}
\begin{aligned}
L_{align} =  \sum_{i=1}^B \{  & \log[1+e^{-\tau_p(\mathbf{S}_i^+-\alpha)}] + \log[1+e^{\tau_n(\mathbf{S}_i^--\beta)}] \\ 
+ & \log[1+e^{-\tau_{MID}(\mathbf{S}_i^{MID+}-\gamma)}]\},
\end{aligned}
\end{equation}
where $\{\tau_p, \tau_n, \tau_{MID}\}$ refer to the scale factors towards global positive, global negative, and multi-integrity positive pairs between visual and textual features, $\{\alpha$, $\beta$, $\gamma\}$ refer to a similarity thresholds of these pairs, and $\{\mathbf{S}_i^+, \mathbf{S}_i^-, \mathbf{S}_i^{MID+}\}$ refer to cosine similarities of these pairs. 

On the other hand, if we introduce these multi-integrity descriptions besides the original ones during training, and regard them as positive pairs, we should also restrict the similarity relationship between the original and these broken text features. Taking the original text as the complete description, and the visual features as the anchor, the textual features from the incomplete descriptions should not be more similar to them than the complete description. The integrity ranking loss is calculated as
\begin{equation}
\begin{aligned}
L_{int} = \sum_{i=1}^B\sum_{k=0}^{K_m-1} max\{0, S(v_i, t_{i}^{(k)})-S(v_i, t_i)\},
\end{aligned}
\end{equation}

If we generate no MIDs, \textit{i.e.} $K_m$ is set to 0, the last terms in $L_{cls}$ and $L_{align}$ will have no effects, and the $L_{int}$ will be always 0. MIDC losses will degenerate into ordinary classification and metric losses in TPS tasks.

\begin{figure}[t]
\centering
\subfloat[\label{subfig-4-a}]{%
   \includegraphics[width=1.0\linewidth]{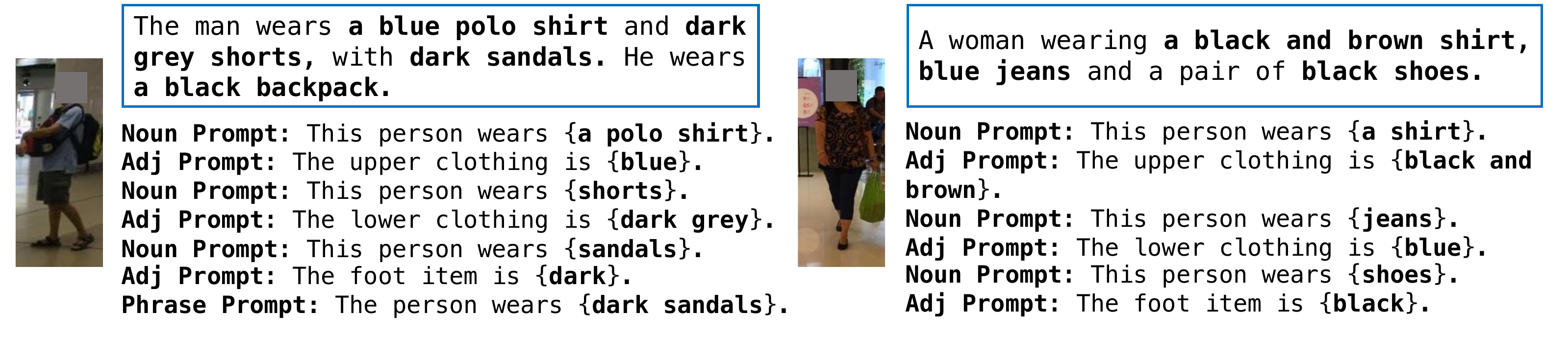}
}

\subfloat[\label{subfig-4-b}]{%
   \includegraphics[width=1.0\linewidth]{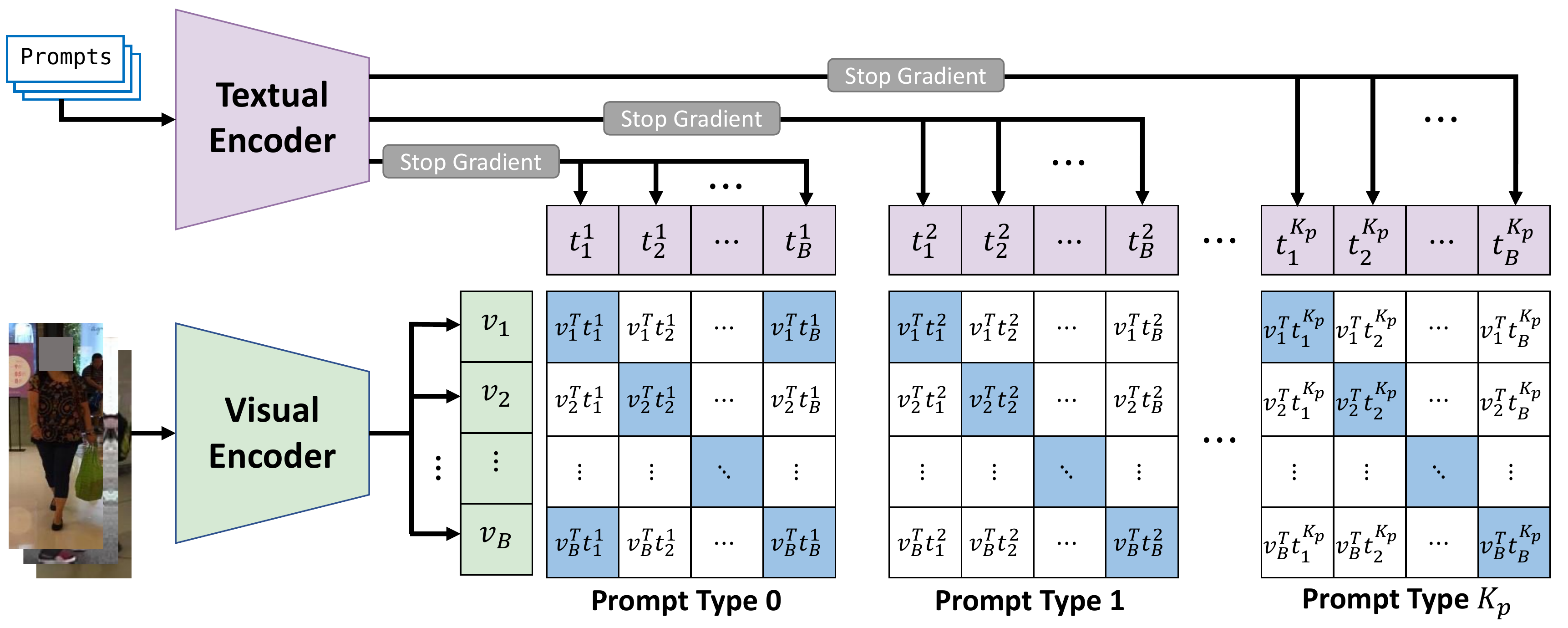}
}

\caption{\label{figure4} Illustration of Dynamic Attribute Prompt: (a) Examples of designed prompt corresponding to a given sentence. (b) Pipeline of DAP. }
\end{figure}

\subsection{Dynamic Attribute Prompt}
VLP models empower not only powerful basic performance for fine-tuning downstream tasks, but also have powerful zero-shot generalization capabilities. Existing methods by fine-tuning on pre-trained models just utilize the former advantage, but the latter is often ignored. Prompt methods are often utilized based on VLP models to achieve the zero-shot classification task \cite{radford2021learning}, $i.e.$ given a template prompt ($e.g.$ “it is \underline{\quad}.”, “a photo of \underline{\quad}.”, “this is a \underline{\quad}.”) prepended to all the candidate classes to form complete text descriptions, and calculate the image-text similarities to get the most similar description as the prediction. The text descriptions in TPS are structural to construct a similar prompt text corpus. Inspired by this motivation, we introduce the Dynamic Attribute Prompt (DAP) task in our method. Figure \ref{figure4} shows an overview of DAP. We also extract all the attribute phrases in each description, but oppositely we construct new descriptions of single attributes with 3 types of prompts, as shown in Figure \ref{subfig-4-a}:

\textbf{1) Prompt for nouns.~} Noun phrases are core stylish descriptions of coarse-grained information in TPS, of which the noun can represent the type of  appearance, decorations or accessories that a person wears or carries, or some gender information. We design the prompt sentences as follows:
\begin{equation}
T_{pmt}=\left\{
    \begin{aligned}
    & \text{“This person is a $\{gender\_nouns\}$.”} \\
    & \text{“This person wears $\{wearing\_nouns\}$.”} \\
    & \text{“This person has $\{decorations\_nouns\}$.”},
    \end{aligned}
\right.
\end{equation}

The semantic nouns can be substitute to exact nouns corresponding to the given person image. Prompt design is very important that using the same sentence prompt indiscriminately can make the generated description syntactically confusing, so we use different predicate words to 3 kinds of noun prompts for grammar correctness.

\textbf{2) Prompt for adjectives.~} Adjectives give fine-grained information in noun phrases. In most cases, they can be color words ($e.g.$ red, blue, green...), or some detail shape modifier about the nouns ($e.g.$ short, round, squared...). The prompt for adjectives can be formulated as:
\begin{equation}
T_{pmt}= \text{“The \{attr item\} of this person is $\{ adjectives \}$.”},
\end{equation}

Considering subsequent computation costs, the subject of this prompt cannot be exact nouns with explicit characteristics. To fill the ``attr item'' blank, we summarize all the attributes nouns into 5 attribute items: upper clothing, lower clothing, head item, foot item, and accessory, corresponding to 5 kinds of adjective prompts.

\textbf{3) Prompt for attribute phrases.~} We can also combine the two kinds of prompts that the prompt template is the same as prompts for nouns, but fill the blank with the whole noun phrases. We can regard this prompt as the reverse operation of generating a MID, $i.e.$ only maintain a certain attribute from the original description. 

The task pipeline of DAP is shown in Figure \ref{subfig-4-b}. For each input image-text pair during fine-tuning, we generate $K_p$ prompts from all kinds of possible prompts. To ensure semantic consistency, we only consider the metric relationship within the same prompt pairs. The dynamic attribute prompt loss can be calculated as
\begin{equation}
\begin{aligned}
    L_{pmt} =  \sum_{i=1}^B\sum_{k=1}^{K_{p}}[&\sum_{t_{pmt}^+ \in \mathcal{P}^i_{(k)}}{\log[1+e^{-\tau_p(S(v_i, t_{pmt}^+)-\alpha)}]} + \\
    & \sum_{t_{pmt}^- \in \mathcal{N}^i_{(k)}}{\log[1+e^{\tau_n(S(v_i, t_{pmt}^-)-\beta)}]}],
\end{aligned}
\end{equation}
where the parameters $\alpha$ and $\beta$ is the same as MIDC. The weights of our textual encoders are not frozen to adapt the certain text structure of TPS. Considering the prompt text contents are not totally aligned to the given text descriptions, the optimization on the textual encoder should not be effected by these prompts. So we conduct back-propagation only to the visual encoder, which means DAP can reversely guide the visual modality to enhance the fine-grained representation learning.

Within each kind of prompt, we intuitively denote the corresponding image-prompt pairs and intra-ID image-text pairs as positive, and the others as negative. However, a prompt description might match the appearances of inter-ID persons, or mismatch intra-ID persons in some cases. To avoid these cases, we utilize the side information from text-wise similarities to correct the pos-neg relationship as 
\begin{equation}
\begin{aligned}
    &\mathcal{P}^i_{(k)} = \{t_{pmt(k)}^j | (y_j = y_i) \cup (S(t_{pmt(k)}^j, t_{pmt(k)}^i) \ge th) \} \\
    &\mathcal{N}^i_{(k)} = \{t_{pmt(k)}^j | (y_j \ne y_i) \cap (S(t_{pmt(k)}^j, t_{pmt(k)}^i) < th) \},
\end{aligned}
\end{equation}

The total loss of TP-TPS training pipeline can be summarized as
\begin{equation}
L = L_{cls} + L_{align} + \lambda_0 L_{int} +\lambda_1 L_{pmt}.
\end{equation}

\section{Experiments}
\subsection{Datasets and Protocols}
Our experiments are conducted on three mainstream benchmark datasets for TPS. CUHK-PEDES \cite{li2017person} is the first and most popular evaluation benchmark for this task. Its training set consists of 34,054 images 11,003 person IDs, and 2 corresponding sentence captions for each image, and its validation and test sets respectively consist of 3,078 and 3,074 images of 1,000 IDs. ICFG-PEDES \cite{ding2021semantically} is an intra-source image-text dataset with challenging MSMT17 database \cite{wei2018person} for person reID task, consists of 34,674 pairs of images and corresponding captions of 3,102 person IDs in the training set, and 19,848 pairs of 1,000 IDs in the test set. RSTPReid \cite{zhu2021dssl} is constructed for real scenarios, also based on MSMT17. contains 20,505 images of 4,101 IDs for all training, validation, and test sets. 

We evaluate the commonly used top-\textit{k} accuracy on all the benchmarks. A query sentence caption is given to retrieve its corresponding person in the image galleries. The top-\textit{k} person search is successful if the corresponding person appears in the top-{k} rankings according to image-text similarities. We evaluate the top-1, 5, 10 accuracies in our experiments. Besides, some works also report mAP metrics to evaluate the ranking performance in detail, so we also report this according to the person reID style.

\begin{table}
\centering
\caption{\label{table1} Comparison of state-of-the-art methods on CUHK-PEDES dataset. $
^\#$: Textual encoder is initialized with BERT pre-trained model. $^\dag$: Visual and textual encoder is partially or wholly initialized with CLIP pre-trained model.}
\begin{tabular}{c|cccc} 
\hline
Methods              & Rank1           & Rank5           & Rank10          & mAP              \\ 
\hline
GNA-RNN \cite{li2017person}             & 19.05~          & -               & 53.64~          & -                \\
CMPM/C \cite{zhang2018deep}              & 49.37~          & -               & 79.27~          & -                \\
TIMAM \cite{sarafianos2019adversarial}               & 54.51~           & 77.56~          & 84.78~          & -                \\
ViTAA \cite{wang2020vitaa}               & 55.97~          & 75.84~          & 83.52~          & -                \\
HGAN \cite{zheng2020hierarchical}                 & 59.00~          & 79.49~          & 86.60~          & -                \\
DSSL \cite{zhu2021dssl}                & 59.98~          & 80.41~          & 87.56~          & -                \\
MGEL \cite{wang2021text}               & 60.27~          & 80.01~          & 86.74~          & -                \\
SSAN \cite{ding2021semantically}                & 61.37~          & 80.15~          & 86.73~          & -                \\
NAFS$^\#$ \cite{gao2021contextual}          & 61.50~          & 81.19~          & 87.51~          & -                \\
AXM-Net$^\#$ \cite{farooq2022axm}             & 61.90~          & 79.41~          & 85.75~          & 57.38~           \\
\textbf{CLIP+BERT$^\#$} & \underline{62.91}~ & \underline{80.99}~ & \underline{86.89}~ & \underline{60.27}~ \\
TIPCB$^\#$ \cite{chen2022tipcb}               & 63.63~          & 82.82~          & 89.01~          & 56.78~           \\
LapsCore$^\#$ \cite{wu2021lapscore}           & 63.40~          & -               & 87.80~          & -                \\
ISANet \cite{yan2022image}              & 63.92~          & 82.15~          & 87.69~          & -                \\
TBPS-LD$^\dag$\cite{han2021text}             & 64.08~          & 81.73~          & 88.19~          & 60.08~           \\
LGUR$^\#$ \cite{shao2022learning}          & 64.21~          & 81.94~          & 87.93~          & - \\
IVT$^\#$\cite{shu2022see}                 & 65.59~          & 83.11~          & 89.21~          & 60.66~                \\ 
\hline
\textbf{VLP-TPS$^\dag$} & 65.38~ & 82.74~ & 88.98~ & 62.47~ \\
\textbf{TP-TPS$^\dag$} & \textbf{70.16~} & \textbf{86.10~} & \textbf{90.98~} & \textbf{66.32~}  \\
\hline
\end{tabular}
\end{table}

\subsection{Implementation}
We use the ViT-B architecture \cite{dosovitskiy2020image} as the visual backbone, and a 12-layer stacked transformers as the textual backbone. The output embedding dimension is set to 512. All the backbones are initialized by CLIP pre-trained weights from OpenAI \cite{radford2021learning}. The input images are all resized into 384 $\times$ 128. We apply horizontal flipping, random cropping and random erasing \cite{zhong2020random} as data augmentations. During training, besides the image-text pair sampling, we generate $K_m=3$ multi-integrity descriptions and $K_p=3$ attribute prompts by open-sourced NLP-toolkits SpaCy. The loss parameters are set with $s=30.0$, $\{\alpha, \beta, \gamma\}=\{0.6, 0.4, 0.6\}$, and $\{\tau_p, \tau_n, \tau_{MID}\}=\{10.0, 40.0, 15.0\}$.  We set loss factor $\lambda_0=0.0001$ and $\lambda_1=0.01$. The threshold of DAP positive pairs is set to 0.98. Similar to common transformer fine-tuning pipelines, AdamW optimizer \cite{loshchilov2017decoupled} is used during training, with the initial learning rate 1e-5 and the weight decay factor 0.1. A cosine learning rate decay schedule is applied after 5 epochs linear warm-up training. The total training epoch lasts for 60 epochs. All the training and inference processes are accelerated on 4 NVIDIA A100 GPUs, implemented with PyTorch framework. 

\begin{table}
\centering
\caption{\label{table2} Comparison of state-of-the-art methods on ICFG-PEDES.}
\begin{tabular}{c|cccc} 
\hline
Methods     & R1     & R5     & R10    & mAP     \\ 
\hline
Dual Path \cite{zheng2020dual}  & 38.99~  & 59.44~  & 68.41~  & -       \\
CMPC/C \cite{zhang2018deep}     & 43.51~ & 65.44~ & 74.26~ & -       \\
MIA \cite{niu2020improving}         & 46.49~ & 67.14~ & 75.18~ & -       \\
SCAN \cite{lee2018stacked}       & 50.05~ & 69.65~ & 77.21~ & -       \\
ViTAA \cite{wang2020vitaa}      & 50.98~ & 68.79~ & 75.78~ & -       \\
\textbf{CLIP+BERT} & \underline{53.98}~ & \underline{72.21}~ & \underline{79.13}~ & \underline{38.62}~ \\
SSAN \cite{ding2021semantically}       & 54.23~ & 72.63~ & 79.53~ & -       \\
IVT \cite{shu2022see}        & 56.04~ & 73.60~ & 80.22~ & -       \\
LGUR \cite{shao2022learning}    & 57.42~ & 74.97~ & 81.45~ & -       \\ 
ISANet \cite{yan2022image}     & 57.73~ & 75.42~ & 81.72~ & -       \\ 
\hline
\textbf{VLP-TPS} & 56.79~ & 72.70~ & 78.98~ & 40.59~  \\
\textbf{TP-TPS} & \textbf{60.64~}  & \textbf{75.97~} & \textbf{81.76~} & \textbf{42.78~}  \\
\hline
\end{tabular}
\end{table}

\begin{table}
\centering
\caption{\label{table3} Comparison of state-of-the-art methods on RSTPReid.}
\begin{tabular}{c|cccc} 
\hline
Method               & R1     & R5     & R10    & mAP     \\ 
\hline
DSSL \cite{zhu2021dssl} & 32.43~ & 55.08~ & 63.19~  & -       \\
SSAN \cite{ding2021semantically} & 43.50~ & 67.80~ & 77.15~  & -       \\
\textbf{CLIP+BERT} & \underline{44.75}~ & \underline{68.25}~ & \underline{77.85}~ & \underline{38.01}~ \\
LBUL \cite{wang2022look} & 45.55~ & 68.20~ & 77.85~  & -       \\
IVT \cite{shu2022see} & 46.70~ & 70.00~ & 78.80~  & -       \\ 
\hline
\textbf{VLP-TPS} & 45.55~ & 68.85~ & 77.60~ & 40.99~  \\
\textbf{TP-TPS} & \textbf{50.65~}  & \textbf{72.45~} & \textbf{81.20~} & \textbf{43.11~}  \\
\hline
\end{tabular}
\end{table}

\subsection{Comparison with State-of-the-Art methods}
\textbf{CUHK-PEDES.~} As shown in Table \ref{table1}, we compare the proposed VLP- and TP-TPS models with other state-of-the-art methods. Our proposed TP-TPS model achieves 70.16\% R1 accuracy and 66.32\% mAP, surpassing the previous best model IVT by a large margin of 4.47\% R1 and 5.66\% mAP. As far as we know, this is the first method that achieves higher than 70\% R1 accuracy among all the TPS methods. This performance achieves by joint three aspects: 1) The bonus from VLP models. Naive VLP-TPS surpasses most of the previous models. 2) Pre-trained textual encoders. VLP-TPS achieves a similar baseline initialized with only CLIP visual encoder (CLIP+BERT) by 2.47\% Rank-1 accuracy and 1.80\% mAP. 3) Proposed MIDC and DAP modules exploit the textual potential of CLIP, make TP-TPS outperform the baseline by 4.78\% R1 and 3.85\% mAP. 

\textbf{ICFG-PEDES.~} Comparison is shown in Table \ref{table2}. Our proposed TP-TPS method achieves 60.64\% R1 accuracy and 42.78\% mAP on the ICFG-PEDES dataset. TP-TPS outperforms the dataset baseline SSAN by 6.41\% R1. Compared to ISANet \cite{yan2022image}, TP-TPS do not consists of complex operations or constraints, but surpasses by 3.91\% R1 accuracy, which shows the priority of the TP-TPS model. Compared to our baseline VLP-TPS, TP-TPS surpasses by 4.85\% R1 and 2.19\% mAP, which verifies the effectiveness of our proposed MIDC and DAP modules on this dataset.

\textbf{RSTPReid.~} RSTPReid is a recent benchmark that focuses on real scenarios. According to the comparison result in Table \ref{table3}, our proposed TP-TPS model also performs well with 50.65\% R1 accuracy and 43.11\% mAP. Comparing the baseline VLP-TPS, TP-TPS surpasses 5.10\% and 2.12\% on R1 and mAP. Comparing TP-TPS outperforms the SoTA IVT model \cite{shu2022see} by 4.95\% R1 accuracy, and the dataset baseline DSSL \cite{zhu2021dssl} by 18.22\% R1 accuracy. 
\begin{table}[t]
\caption{\label{table4} Ablation results of all components evaluated on CUHK-PEDES dataset. ``CLIP-TE'' refers to initialized with the pre-trained textual encoder of CLIP model. ``MIDC'' refers to generating no less than one MIDs and applying the corresponding constriant losses. ``DAP'' refers to generating no less than one kind of prompt and applying the corresponding loss.}
\centering
\begin{tabular}{c|cccc|cccc}
\hline
No. & BL                        & CLIP-TE                      & MIDC                      & DAP                       & R1                              & mAP                             \\ \hline
1   & \checkmark &                           &                           &                           & 62.91                           & 60.27                           \\
2   & \checkmark &                           & \checkmark &                           & 65.43                           & 63.00                           \\
2   & \checkmark &                           &                           & \checkmark & 61.40                           & 59.02                           \\ \hline
3   & \checkmark & \checkmark &                           &                           & 65.38                           & 62.47                           \\
4   & \checkmark & \checkmark & \checkmark &                           & 68.73                           & 64.72                           \\
5   & \checkmark & \checkmark &                           & \checkmark & 68.40                           & 64.04                           \\
6   & \checkmark & \checkmark & \checkmark & \checkmark & \textbf{70.16} & \textbf{66.32} \\ 
\hline
\end{tabular}
\end{table}

\begin{figure}[t]
\centering
\subfloat[\label{subfig-5-a} $K_m$]{%
   \includegraphics[width=0.5\linewidth]{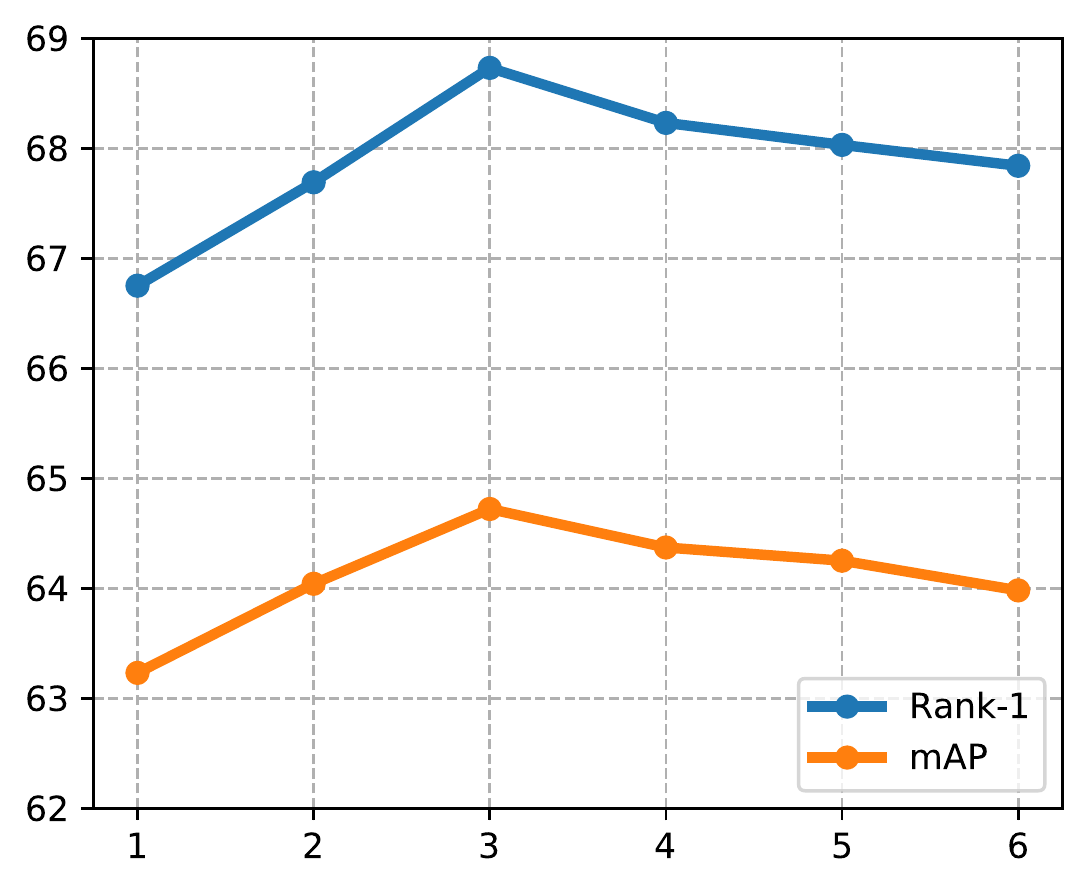}
}
\subfloat[\label{subfig-5-b} $K_p$]{%
   \includegraphics[width=0.5\linewidth]{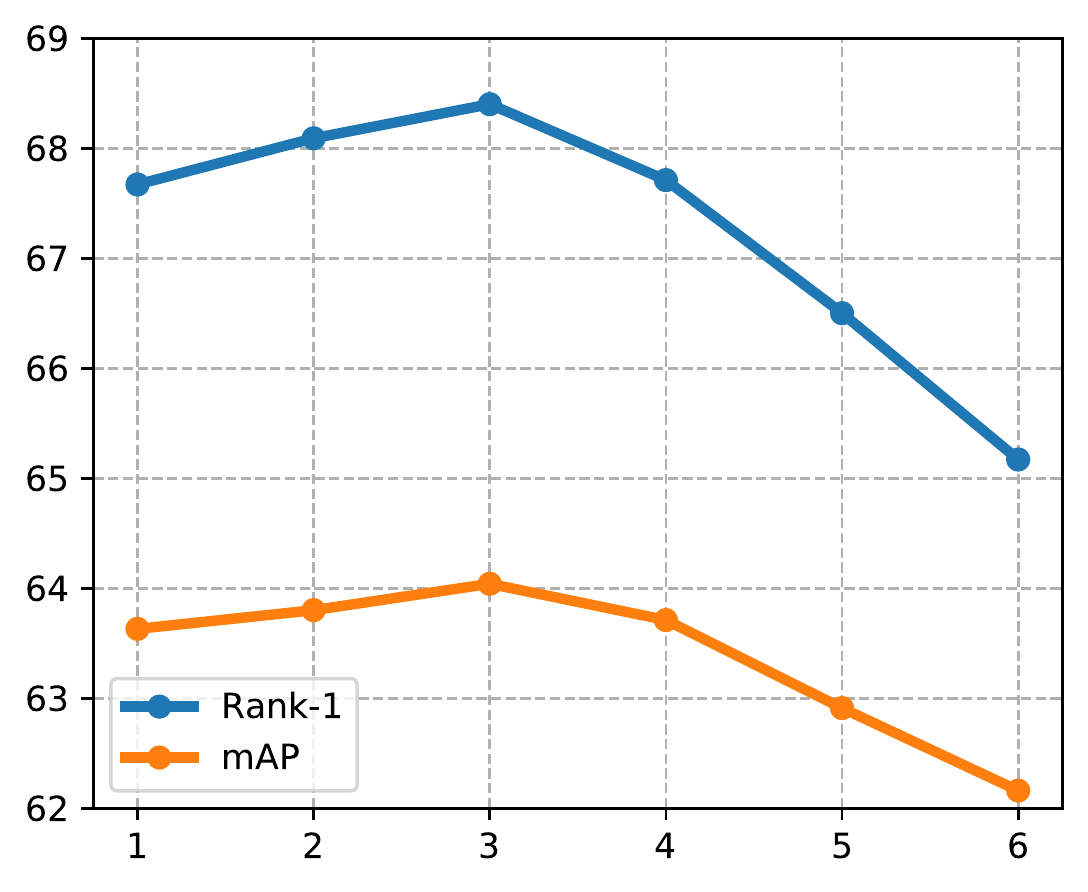}
   
}
\caption{Effects of Parameters, including (a) the number of MID $K_m$, and (b) the number of different prompts $K_p$.}
\end{figure}

\begin{figure*}
\centering
\includegraphics[width=1.0\linewidth]{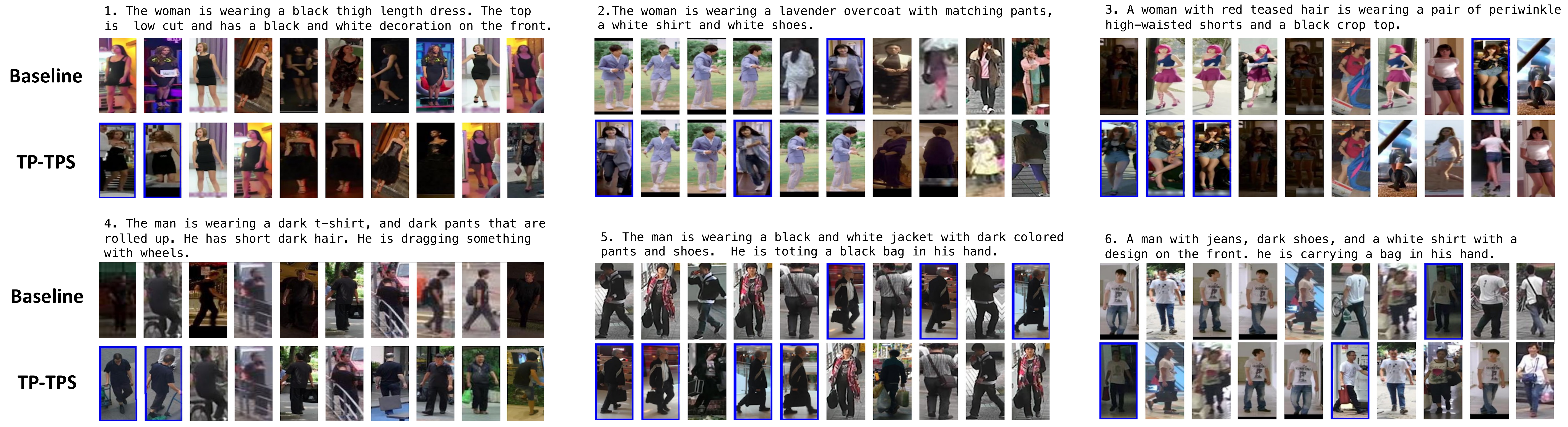}
\caption{\label{figure6} Visualization of Top-10 ranking results by the baseline and TP-TPS models for 6 query sentences. The image with blue borders are ground-truth matching person image.}
\end{figure*}

\subsection{Ablation Study}
Some major ablation study results are demonstrated in this section. Restricted by the page limits, some more results are also given in the supplemental materials.

\textbf{Effectiveness of MIDC.~} Table \ref{table4} gives the performance comparison between models employed with and without MIDC. For models without CLIP textual encoder, MIDC pushs the performance from 62.91\% R1 and 60.27\% mAP to 65.43\% R1 and 63.00\% mAP, improved by 2.52\% R1 and 2.73\% mAP. For models with CLIP-TE, MIDC improves from 65.38\% R1 and 62.47\% mAP to 68.73\% R1 and 64.72\% mAP, improved by 3.35\% R1 and 2.25\% mAP. We can observe that MIDC brings about 2\% larger improvement based on a better baseline model. The performance gains from CLIP-TE do not overlap or conflict with MIDC, and have a mutually reinforcing effect. The better discrimination capability by textual transferable knowledge of CLIP makes the textual tolerance with different integrity.

\textbf{Effectiveness of DAP.~}Table \ref{table4} shows the effectiveness brought by DAP. We get some seemingly contradictory but reasonable results. Based on CLIP-TE, DAP achieves 68.40\% R1 accuracy and 64.04\% mAP, surpassing the baseline by 3.02\% R1 and 1.57\% mAP. Working with MIDC, DAP pushes R1 and mAP to 70.16\%  and 66.32\% respectively, surpassing by 1.24\% R1 and 2.28\% mAP. Overlap of performance gains between two modules can be observed. Considering the pipeline of MIDC, the description without some adjectives can be regarded as a non-unified noun prompt, that has similar effects to DAP. On the other hand, based on models without CLIP-TE, DAP brings slight performance degradation by 1.51\% R1 and 1.25\% mAP. CLIP-TE established a relationship with text descriptions and image contents, but using an external language model might break this correlation, which makes attributes prompts not helpful as a textual hint of visual fine-grained details. 

\textbf{Effectiveness of VLP Textual Encoders.~} As shown in Table \ref{table4}, the baseline backbone without CLIP-TE can achieve 62.91\% R1 accuracy and 60.27\% mAP. Initialization with pre-trained TE improves 2.47\% R1 and 2.20\% mAP, achieving 65.38\% R1 and 62.47\% mAP. This result reflects the positive effects of textual transferable knowledge on accuracy performance. Unexpected bonus after employing the MIDC and DAP further verifies the effectiveness of dual pre-training.

\textbf{Parameter Analysis on Number of MID $K_m$.~} The effect of different $K_m$ settings is illustrated in Figure \ref{subfig-5-a}. The range of performance varies from 66.75\% to 68.73\% on R1, and 63.23\% to 64.72\% on mAP. The setting with $K_m=3$ achieves the best performance. Even with only one additional MID ($K_m=1$), MIDC can still achieve 1.37\% R1 and 0.76\% mAP improvement. This reveals MIDC is a robust constraint for better generalization. Improvement by a larger $K_m$ configures is smoothly smaller, but introduce more memory burden, which are not efficient options. 

\textbf{Parameter Analysis on Number of Prompts $K_p$.~} Figure \ref{subfig-5-b} shows the effects of different $K_p$ settings. The performance ranges from 65.17\% to 68.40\% on R1, and 62.16\% to 64.04\% on mAP. As the same as MIDC, $K_p=3$ is the best configure, and $K_p=1$ can achieve similar performance to $K_p=3$. However, when $K_p$ comes to larger than 4, a drastic drop can be observed. $K_p=6$ even brings the model worse than the baseline. According to statistics, the mean number of structural attributes of CUHK-PEDES decriptions comes to about 2 to 3. During training, if no enough available prompts can be generated, the existing prompts will be sampled. Besides, restricted by the performance of syntax analysis, the quality of prompts might have bad affects on this result.

\subsection{Visualization}
Figure \ref{figure6} shows some ranking results obviously improved by TP-TPS, compared with the baseline model. We can observe an intuitive effects by MIDC and DAP from some rare details in the dataset distribution. In Result 1, TP-TPS comprehensively represents the stylish description ``low cut'',  decorations on clothing ``black and white decoration on the front''. Such fine-grained information is ignored by the baseline. In Result 3, short skirts and shorts are accurately distinguished by TP-TPS, as well as red hair and pink hair. Such synonym descriptions are garbled by the baseline model. In Result 4, 5 and 6, accessory items such as ``something with wheels'', ``toting black bag in his hand'' and ``carrying a bag in his hand'' are concentrated by TP-TPS. It shows a better association effects comparing to models without such constraints. 

\section{Conclusion}
In this paper, we explore the potential of text-based person search models with vision-language pre-training by proposing the TP-TPS framework. To this end, we introduce a simple baseline model called VLP-TPS, which efficiently fine-tunes dual pre-trained encoders. Moreover, we present two key components to enhance the textual potential of the model. Firstly, we incorporate the Multi-Integrity Description Constraints (MIDC) to ensure the consistency between generated fine-grained missing sentences and the original image-text pairs. Secondly, we introduce the Dynamic Attribute Prompt (DAP) to guide image representations with fine-grained attributes of natural language descriptions. Extensive experiments demonstrate the remarkable performance of TP-TPS.

{\small
\bibliographystyle{ieee_fullname}
\bibliography{egbib}
}

\end{document}